\documentclass[conference]{IEEEtran}

\usepackage{graphicx}
\usepackage{amsmath}
\usepackage{amssymb}
\usepackage{booktabs}

\usepackage{amsfonts}
\usepackage[noend]{algorithmic}
\usepackage{algorithm}

\usepackage{textcomp}
\usepackage{xcolor}
\usepackage{multirow}
\usepackage{siunitx}
\DeclareSIUnit\pixel{px}
\usepackage{bbm, dsfont} 
\usepackage[normalem]{ulem}
\usepackage{subfig}
\usepackage{url}

\usepackage{forest}

\definecolor{folderbg}{RGB}{124,166,198}
\definecolor{folderborder}{RGB}{110,144,169}

\def\Size{4pt}
\tikzset{
  folder/.pic={
    \filldraw[draw=folderborder,top color=folderbg!50,bottom color=folderbg]
      (-1.05*\Size,0.2\Size+5pt) rectangle ++(.75*\Size,-0.2\Size-5pt);  
    \filldraw[draw=folderborder,top color=folderbg!50,bottom color=folderbg]
      (-1.15*\Size,-\Size) rectangle (1.15*\Size,\Size);
  }
}

\usepackage[numbers]{natbib}

\usepackage[hidelinks]{hyperref}
\usepackage[capitalize]{cleveref}

\newcommand{\eg}{e.g.\,}

\usepackage{tikz}

\newcommand\submittedtext{%
  \footnotesize \textcopyright \the\year{} IEEE. Appears in \textit{2025 IEEE 12th International Conference on Data Science and Advanced Analytics (DSAA)}, \\ DOI: 10.1109/DSAA65442.2025.11248012}

\newcommand\submittednotice{%
\begin{tikzpicture}[remember picture,overlay]
\node[anchor=south,yshift=10pt] at (current page.south) {\parbox{\dimexpr0.8\textwidth-\fboxsep-\fboxrule\relax}{\submittedtext}};
\end{tikzpicture}%
}

\newcommand\copyrighttext{%
  \footnotesize \textcopyright \the\year{} IEEE. Personal use of this material is permitted. Permission from IEEE must be obtained for all other uses, in any current or future media, including reprinting/republishing this material for advertising or promotional purposes, creating new collective works, for resale or redistribution to servers or lists, or reuse of any copyrighted component of this work in other works.}

\newcommand\copyrightnotice{%
\begin{tikzpicture}[remember picture,overlay]
\node[anchor=north,yshift=-10pt] at (current page.north) {\parbox{\dimexpr1\textwidth-\fboxsep-\fboxrule\relax}{\copyrighttext}};
\end{tikzpicture}%
}

\begin{document}

\title{From Engineering Diagrams to Graphs: \\ Digitizing P\&IDs with Transformers
}

\author{\IEEEauthorblockN{1\textsuperscript{st} Stürmer, Jan Marius}
\IEEEauthorblockA{\textit{German Aerospace Center (DLR)} \\
\textit{Institute for the Protection of}\\
\textit{Terrestrial Infrastructures}\\
Sankt Augustin, Germany \\
jan.stuermer@dlr.de}
\and
\IEEEauthorblockN{2\textsuperscript{nd} Graumann, Marius}
\IEEEauthorblockA{\textit{German Aerospace Center (DLR)} \\
\textit{Institute for the Protection of}\\
\textit{Terrestrial Infrastructures}\\
Sankt Augustin, Germany \\
marius.graumann@dlr.de}
\and
\IEEEauthorblockN{3\textsuperscript{rd} Koch, Tobias}
\IEEEauthorblockA{\textit{German Aerospace Center (DLR)} \\
\textit{Institute for the Protection of}\\
\textit{Terrestrial Infrastructures}\\
Sankt Augustin, Germany \\
tobias.koch@dlr.de}
}

\maketitle

\copyrightnotice
\submittednotice

\begin{abstract}
Digitizing engineering diagrams like Piping and Instrumentation Diagrams (P\&IDs) plays a vital role in maintainability and operational efficiency of process and hydraulic systems.
Previous methods typically decompose the task into separate steps such as symbol detection and line detection, which can limit their ability to capture the structure in these diagrams.
In this work, a transformer-based approach leveraging the Relationformer that addresses this limitation by jointly extracting symbols and their interconnections from P\&IDs is introduced.
To evaluate our approach and compare it to a modular digitization approach, we present the first publicly accessible benchmark dataset for P\&ID digitization, annotated with graph-level ground truth.
Experimental results on real-world diagrams show that our method significantly outperforms the modular baseline, achieving over 25\% improvement in edge detection accuracy. 
This research contributes a reproducible evaluation framework and demonstrates the effectiveness of transformer models for structural understanding of complex engineering diagrams. 
The dataset is available under \url{https://zenodo.org/records/14803338}.
\end{abstract}


\section{Introduction}
\label{sec:introduction}

The design process of complex technical systems starts with a conceptual engineering drawing that describes the properties of hydraulic components and instrumentation as well as their interconnections. 
These properties and relationships can be represented by an attributed, directed graph or relational structured data formats. 
Machine-readable information about technical systems can be used to support planning and operation of these systems through simulation using state-of-the-art frameworks.
Nowadays, such plans are created digitally using CAD software and stored as machine-readable data. 
However, there is a large old stock of such plans in non-digital form. 
Furthermore, the exchange of such plans often takes place via rendered images or PDF files.
This is linked to a rising demand for digitization of such documents.
One type of diagrams are Piping and Instrumentation Diagrams (P\&ID).
A P\&ID is a detailed diagram used in the chemical process industry, which describes the equipment installed in a plant, along with instrumentation, controls, piping etc. and is used during planning, operation and maintenance \cite{Toghraei.2019}.

This paper presents several contributions to the field of P\&ID digitization.
Firstly, we describe methods for digitizing P\&IDs based on a recent transformer network along with the modular digitization approach based on previous work.
We describe the metrics used for evaluating the performance of our proposed method, providing a comprehensive framework for assessing its effectiveness.
We then compare our proposed methods using synthetically created and real-world engineering diagrams.
To facilitate further research and development, we publish the small test dataset \textit{PID2Graph}. 
Notably, this dataset is, to the best of our knowledge, the first publicly available P\&ID dataset that contains real-world data and annotations for the full graph, including symbols and their connections.

\section{Related Work}
\label{sec:relwork}

Engineering diagram digitization primarily involves two key tasks: extracting components and their interconnections. 
The successful digitization of P\&IDs enables the automated creation of accurate simulation models, as demonstrated by previous research. For instance, \cite{Paganus.2018} generated models from text-based descriptions, while \cite{Stuermer.2023} developed a pipeline for generating simulation models directly from digitized engineering diagrams.
P\&ID digitization can be approached in two ways: as a series of separate sub-problems solved by one module per sub-problem or as an image-to-graph problem.

\subsection{Modular Engineering Diagram Digitization}
The most commonly employed method to digitize P\&IDs is utilizing separate deep learning models or algorithms for symbol detection, text detection, and line detection.
The connection detection usually is not evaluated.
This approach was first proposed by \cite{Rahul.2019} and has since been refined in subsequent work \cite{Mani.2020, Gada.2021}.
Here, Convolutional Neural Networks (CNN) are employed for symbol and text detections. 
Afterwards, probabilistic hough transform or other traditional computer vision algorithms are applied to detect lines. 
To combine these detections, a graph describing the components and their interconnections is created.
A recent review by \cite{Jamieson.2024} provides an in-depth analysis of existing literature on deep learning-based engineering diagram digitization.
In contrast to employing CNNs for symbol detection (\eg, \cite{Nurminen.2020, Cha.2019}), alternative approaches have been explored, including the use of segmentation techniques \cite{MorenoGarcia.2020} or Graph Neural Networks (GNNs) \cite{Paliwal.2021b}. 
Additionally, other studies have focused on related but distinct tasks, such as line and flow direction detection \cite{Kim.2022} or line classification \cite{Kim.2023}.

The principles of engineering diagram digitization can be extended to other domains, even if the application domain or diagram type differs from P\&IDs.
A recent study by \cite{Theisen.2023} has demonstrated the digitization of process flow diagrams (PFDs), leveraging a dataset comprising approximately 1,000 images sourced from textbooks.
The connections and graph were extracted by skeletonizing the image. However, the connection extraction was not evaluated due to missing labels for this task.
Other approaches deal with electrical engineering diagrams \cite{Bhanbhro.2023, Yang.2024}, handwritten circuit diagram digitization \cite{Bayer.2023}, mechanical engineering sketches \cite{Bickel.2024} or construction diagrams \cite{Jamieson.2024b}.

\subsection{Image to Graph}
\label{sec:relwork_i2g}

Another possibility to tackle engineering diagram digitization is framing it as an image-to-graph problem.
\cite{He.2020} introduce a road graph extraction algorithm that is able to extract a graph describing a road network from satellite images using neural networks.
A similar problem is described by \cite{Belli.2019}, where a \textit{Generative Graph Transformer} is described to extract graphs from images of road networks.
Although the problem of identifying connections in diagrams is similar to extracting connections from engineering diagrams, the challenges of identifying specific symbols and extracting different types of relationships remain. 
The Scene Graph Generation (SGG) task involves solving these challenges, where objects from an image are extracted and the relationships between them are determined \cite{Zhu.2022}.
One transformer network dealing with SGG is SGTR+ \cite{Li.2024}.
A recent framework combining SGG and road network extraction is the \textit{Relationformer} proposed by \cite{Shit.2022}. 
The Relationformer, an enhancement of deformable DETR (DEtection TRansformer) \cite{Zhu.2020}, combines advanced object detection with relation prediction and is shown to outperform existing methods.
Since its release, the Relationformer has successfully been adapted for further work like crop field extraction \cite{Xia.2024}.
Metrics that have been used to evaluate graph extraction evaluation are manifold.
For the problem of road network extraction from images, the metrics Streetmovers-Distance \cite{Belli.2019}, TOPO \cite{Biagioni.2012, Belli.2019} or APLS Metric \cite{vanEtten.2018} were introduced.
Another commonly used relation metric is the scene graph generation metric $\text{mR@}K$. 
Objects are matched by IOU (or similar metrics) and the (object, relation, object)-tuples are ranked by confidence. 
The $K$ most confident results are then used to calculate the mean recall.
For scene graph generation, this approach with using only the top-$K$ results is used because the annotations are highly incomplete due to the high amount of possible relations. \cite{Li.2022}

\begin{figure*}[bt] 
    \centering
    \includegraphics[trim={0 0 2cm 3.25cm},clip,width=\textwidth]{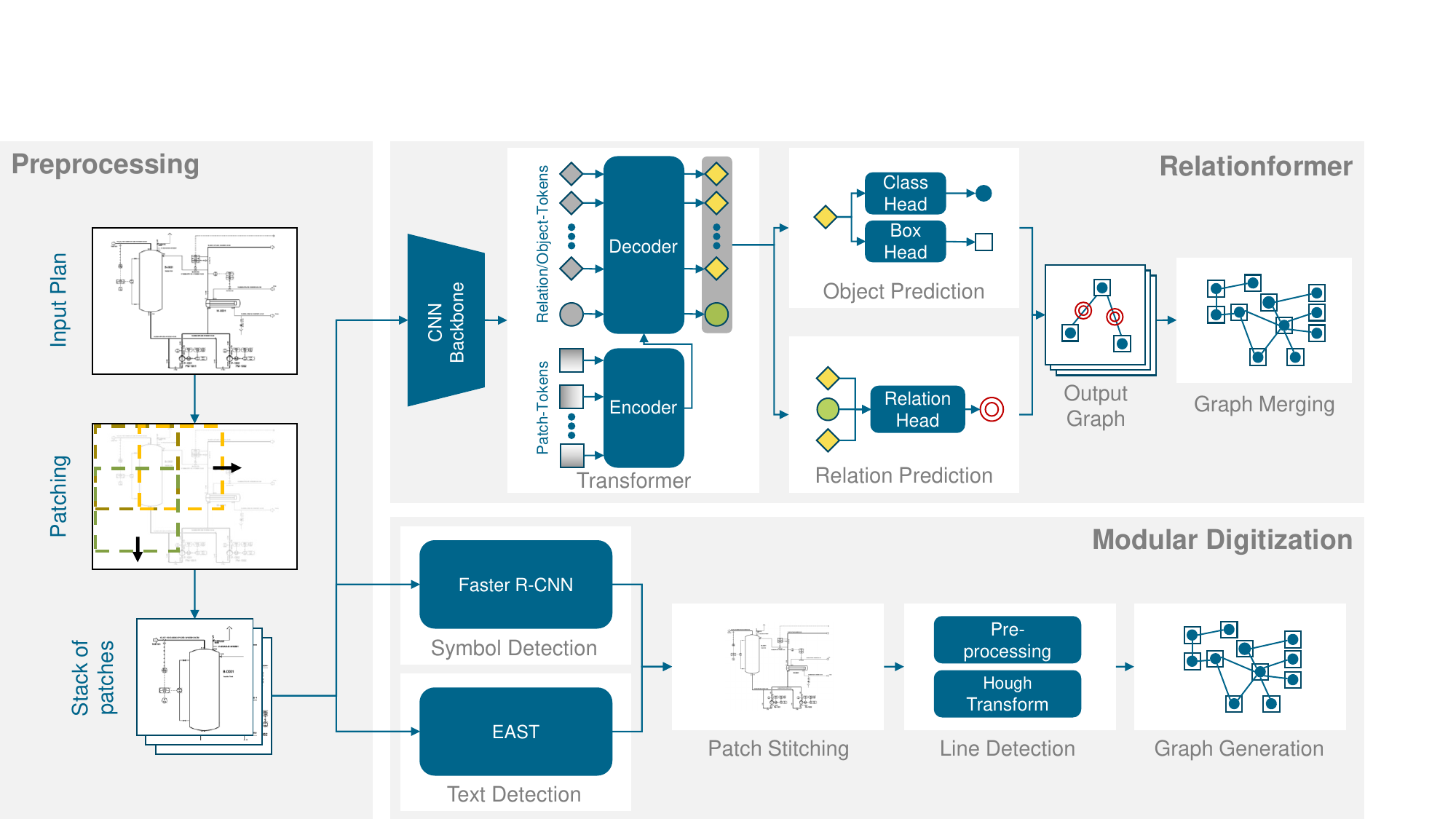}
    \caption{\label{fig:methods}Overview of the Relationformer \cite{Shit.2022} and the Modular Digitization to digitize engineering diagrams. The preprocessing step patches and adjusts the data, which is then fed into each method to produce a graph representation as output.}
\end{figure*}

\subsection{Datasets}

Developing effective models for P\&ID digitization requires P\&IDs that are accurately labeled with symbols and connections. 
Unfortunately, existing research has mostly been evaluated on private datasets, which have not been published due to concerns around copyright and intellectual property rights.
Some approaches use synthetic data \cite{Nurminen.2020, Paliwal.2021} or augment their data applying Generative Adversarial Networks (GANs) \cite{Elyan.2020}.
The only published dataset for P\&ID digitization is Dataset-P\&ID by \cite{Paliwal.2021} consisting of 500 synthetic P\&IDs with 32 different symbol classes.
The data includes annotations for symbols, lines, and text, which are valuable resources. 
However, the graph structure annotations are not provided, and the lines between symbols are drawn using a simplified grid layout. 
Additionally, one lines style can change between dashed and solid forms. 
These limitations may impact the effectiveness of certain models or methods trained or evaluated with this dataset.

\section{Methods}
\label{sec:methods}

Due to its supposed ability to be adaptable, we apply the Relationformer model in the context of engineering diagram digitization, and compare it to a method that breaks down the digitization task into separate sub-tasks of detecting symbols, text, lines, and then extracting graphs, referred to as \textit{Modular Digitization Approach} in this paper.
Both the proposed Relationformer and the Modular Digitization Approach share a common goal of identifying and classifying symbols and lines within P\&IDs. 
Their outputs are unified in the form of a graph representation, where each node contains a bounding box and symbol class, while edges have an edge class.
The models will be trained and evaluated with the classes listed in \cref{tab:classes}, i.e.,\ seven symbol classes and two line classes.
The workflow for the preprocessing step and both methods is visually depicted in \cref{fig:methods} and are discussed in detail in the subsequent sections.

\subsection{Preprocessing}
Due to the high resolution of P\&IDs alongside with big size differences between symbols, the input resolution for the models would have to be high in order to still accurately depict symbols and lines.
We use patching to split the full diagram resized to 4500x7000 into multiple patches with an overlap of at least \SI{50}{\percent}.
During both model processing steps, each patch is independently processed and then combined with its neighbors to form a complete graph representation.

\begin{table}[bt]
\centering
\setlength{\tabcolsep}{2.5pt}
\caption{Classes used for training and evaluation.}
\label{tab:classes}
\begin{tabular}{lll@{\hskip 20pt}l}
\toprule
\multicolumn{3}{c}{Symbols}                   & Lines     \\ \midrule
General     & Pump/Compressor  & Arrow        & Solid     \\
Tank/Vessel & Instrumentation  & Inlet/Outlet & Non-Solid \\
Valve       &                  &              &          \\ \bottomrule
\end{tabular}
\end{table}

\subsection{Relationformer}
\label{sec:relationformer}
The Relationformer is described as in the original paper~\cite{Shit.2022} and implemented in a modified form to adapt to P\&IDs.
The Relationformer is based on deformable DETR \cite{Zhu.2020} and consists of a CNN backbone, a transformer encoder-decoder architecture and heads for object detection and relation prediction (see \cref{fig:methods}).
The main difference to deformable DETR is the decoder architecture and the relation prediction head.
The decoder uses $N+1$ tokens as the first input, where $N$ is the number of object-tokens and a single relation-token.
The second input are the contextualized image features from the encoder.
The objection detection head consists of two components.
Firstly, Fully Connected Networks (FCNs) are employed to predict the location of each object within the image, in form of a bounding box that define the spatial extent of each object. 
Secondly, a single layer classification module is used, assigning a class label to each detected object.
Thus, the output of the object detection head consists of the class and bounding box of the object.
The relation prediction head has a pairwise object-token and a shared relation-token, where a multi layer-perceptron (MLP) predicts the relation $\Tilde{e}^{ij}_{\text{rln}}=\text{MLP}_\text{rln}({o^i,r,o^j}_{i\neq j})$ for every object-token pair $o^i$ and $o^j$ with the relation-token $r$.

\begin{figure}[bt]%
    \centering
    \subfloat[\centering]{{\includegraphics[width=0.23\textwidth]{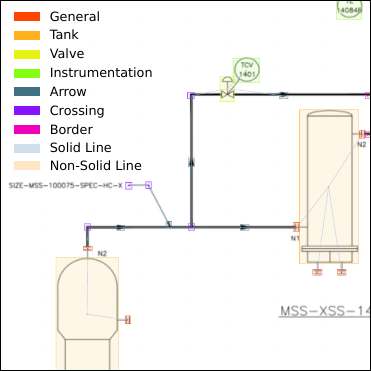}}}%
    \hfill
    \subfloat[\centering]{{\includegraphics[width=0.23\textwidth]{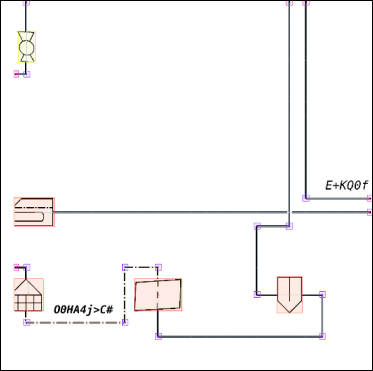}}}%
    \caption{Two example patches obtained after dividing the full diagram of OPEN100 (a) and the Synthetic Test Data (b), with border nodes (pink) and bounding boxes (several colors) marking where lines exit each patch. These patches serve as input to the Relationformer for training, testing and evaluation.}%
    \label{fig:relationformer_input}%
\end{figure}

To adapt the Relationformer model for training on the patched data, several modifications were made to the input graph.
In addition to the pre-existing symbol classes, new node categories were introduced to capture key features of the diagram's structure: line \textit{ankles}, \textit{crossings} and \textit{borders}. 
Nodes get appended to the graph accordingly with a bounding box around the center of the node that has a uniform size, analogously to the procedure described in the original Relationformer paper regarding road networks.
A border bounding box is created when a line gets cut during patching at the position where the line intersects the border of the patch.
Examples for this patched images and graphs can be seen in \cref{fig:relationformer_input}.

Afterwards, the predicted graphs for single patches are merged into a graph representing the complete plan.
The merge process consists of the following steps:
\begin{enumerate}
    \item Lowering the confidence score for each bounding box in a patch with the function $\hat{c}=c-\alpha \cdot e^{-3|d\frac{2}{S}|}$, where $\hat{c}$ is the modified confidence score, $c$ is the original confidence score, $S$ is the size of the patch, $d$ is the smallest distance between the bounding box and the patch border and $\alpha$ is the maximal amount of the weighting $\alpha=0.4$.
    Therefore, symbols with bounding boxes closer to the patch border are more likely to be cut off, as there is a possibility the same symbols is included to a bigger extent in another patch;
    \item Filtering predictions with a low confidence score in order to ignore them during the merging process; 
    \item Collecting and resizing all bounding boxes for the complete plan;
    \item Non-maximum suppression (NMS) with a high IOU threshold to merge duplicates;
    \item Weight-boxed fusion (WBF) with a lower IOU threshold to combine information from bounding boxes;
    \item Cleaning up the graph by removing self-loops and non-connected nodes.
\end{enumerate}

\subsection{Modular Digitization Approach}
An alternate, more commonly used approach involves separately detecting symbols, text and lines and then merging them together into a graph, following earlier work of \cite{Stuermer.2023}, as can be seen in \cref{fig:methods}. 

\paragraph{Symbol Detection}
An improved Faster R-CNN \cite{Li.2021} is used for symbol detection, with patching and merging done as described in \cref{sec:relationformer}.
Because no graph structure is considered by this object detector, the classes for crossings, ankles and border nodes do not exist.

\paragraph{Text Detection}
Text is identified using CRAFT \cite{Baek.2019} and EasyOCR \cite{EasyOCR.2023}.
Notably, this text detection functionality serves as an auxiliary mechanism for filtering purposes only, rather than being evaluated. 
Filtering text prevents the line detection module from incorrectly classifying textual elements as graph connections.

\paragraph{Line Detection}
Lines are detected using dilation, erosion, and Progressive Probabilistic Hough Transform. 
Dashed lines are reconstructed by clustering and merging small line segments.

\paragraph{Graph Generation}
The final step generates a comprehensive graph by assigning line start and end points to symbols or other lines, creating crossing and ankle nodes as needed, and refining the graph by removing self-loops and unused nodes.

\subsection{Datasets}

To address the gap of missing complex and real-world datasets, we have created our own datasets with annotations for symbols and connections.
The synthetic data is generated using symbol templates scrambled and cutout from P\&ID standardization and legends.
Our algorithm randomly places these templates on a canvas and connects them in a way that forms a connected graph. 
If lines intersect, a crossing node is created.
Furthermore, this data includes various line types, such as solid and dashed lines, to enhance the realism of our digitization task and add data diversity.
To further simplify the data and focus on the essential graph structure, additional information such as tables, legends and frames are removed.
The patched training dataset is augmented with various transformations, comprising small angle rotations, \SI{90}{\degree} rotations, horizontal and vertical flips, minor adjustments to brightness and contrast, random scaling, and image blurring.

The training data consist of synthetic and real-world data. \textit{Synthetic 700} is a synthetic dataset with 700 different symbol templates collected from a broad range of sources.
Additionally, we have manually annotated 60 \textit{real-world P\&IDs} from various plants.

For the \textit{PID2Graph Synthetic} dataset, synthetic plans composed exclusively of symbols from \cite{Sandoval.2023}, which are based on the ISO 10628 standard \cite{ISO10628}, are included.

\textit{PID2Graph OPEN100} is used to further validate our method.
It includes 12 manually annotated publicly available P\&IDs from the OPEN100 reactor \cite{open100}. 
Notably, these plans do not contain dashed lines, allowing us to assess the robustness of our model in a scenario where this feature is absent.

To enable a comparison with previous work, we also evaluate our methods on \textit{Dataset-P\&ID} from \cite{Paliwal.2021}. 
This dataset includes annotations in the form of bounding boxes for 32 different symbols and start and end points of line segments, rather than in a graph format. 
We convert these annotations to align with the format used in our other datasets. 
Specifically, we map the 32 symbol classes to our classes as outlined in Table \ref{tab:classes}, and we connect overlapping line segments to create edges. 
Since these edges may include both dashed and solid segments, we assign a unified edge label across all edges.

\begin{figure}[tb] 
  \centering
  \includegraphics[trim={0cm 0cm 0cm 0cm},clip,width=.47\textwidth] {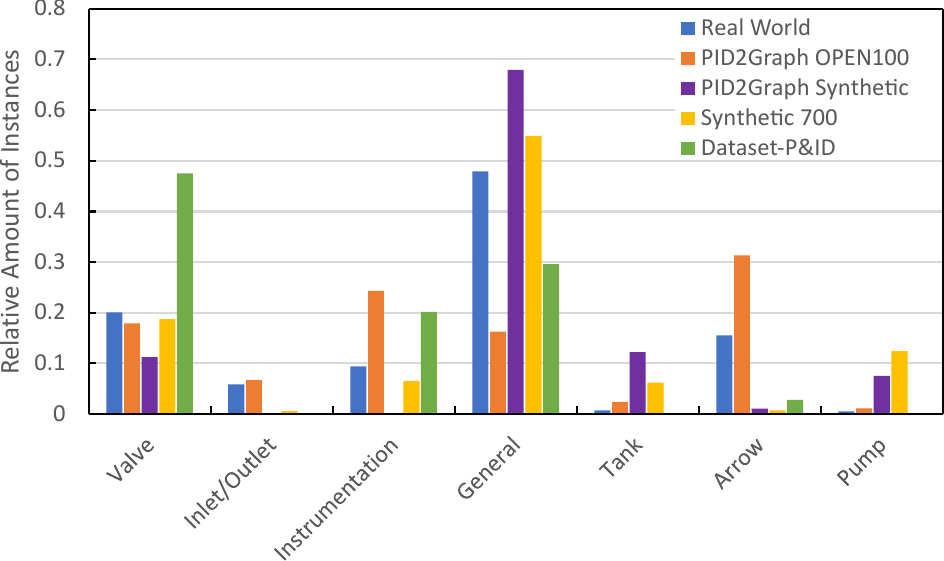}
  \caption{\label{fig:data_stats} Symbol class distribution: Frequency of symbol object classes among the datasets, showing relative abundance of each class type.}
\end{figure}

The relative distribution of symbol classes, excluding crossings and ankles, is depicted in \cref{fig:data_stats}. 
A notable pattern emerges across all datasets: the 'general' class overwhelmingly dominates the distribution in every dataset except OPEN100. 
This can be attributed to the fact that the general class comprises a broad range of symbols, resulting in its disproportionate representation. 

\begin{figure}[tb] 
  \centering
  \includegraphics[trim={0cm 00cm 0cm 0cm},clip,width=.47\textwidth] {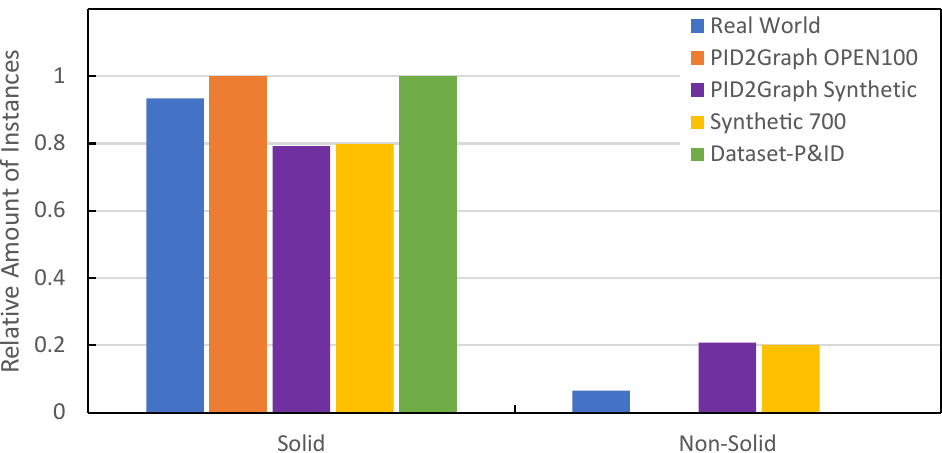}
  \caption{\label{fig:data_stats_edges} Edge Class Distribution: Relative frequency of each edge class across the datasets.}
\end{figure}

The relative distribution of edge classes is shown in \cref{fig:data_stats_edges}. 
The non-solid edge class makes up between 10\% and 20\% of all edges in the datasets except OPEN100.

\subsection{Metrics}
\label{sec:metrics}

We evaluate the quality of graph construction using object detection metrics, where each detected bounding box corresponds to a node in the graph along with a metric for measuring edge detection performance.

\subsubsection{Node Detection}
To comprehensively measure the performance of our method in constructing graph nodes, we employ two metrics.
Firstly, mean Average Precision (mAP) is used to evaluate the estimation of individual symbols, considering only classes from \cref{tab:classes} that do not involve crossings and ankles, as these categories relate to graph connectivity rather than symbol representation.

The second metric used is average precision (AP) across all symbols, crossings and ankles.
However, unlike the first metric, each symbol is assigned to the same class, because an error or a confusion in the assignment of the class to a node is not decisive for the entire graph reconstruction, as long as the node exists. 
This allows us to evaluate the consistency and accuracy of our method in constructing the graphs structure.
Both metrics are calculated using an Intersection over Union (IOU) threshold of \num{0.5}. 
We choose this threshold because exact bounding box localization is not crucial for our application.

\subsubsection{Edge Detection}

\begin{figure}[tb] 
  \centering
  \includegraphics[trim={0 0 26.7cm 8.5cm},clip,width=.4\textwidth]{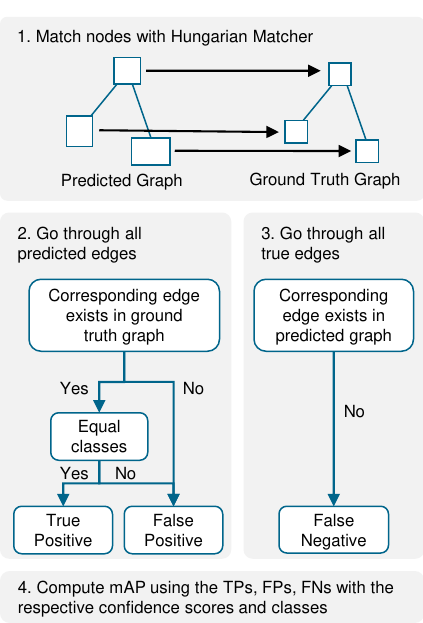}
    \caption{\label{fig:edgemetric_flow} Description of the implemented metric for calculating the edge mAP.}
\end{figure}

Line detection describes the task of detecting the position of a line in the pixel space and it's type, which is used by previous approaches.
In contrast, edge detection is the task of detecting edges between two objects or nodes.
An edge detection metric should reflect the task as independently as possible from node detection.
This is inherently difficult, because an edge is always associated with two nodes.
If one of the nodes is missing, the edge will be missing as well.
For a false positive node, there may be edges connected to that node that would never exist if the node had not been detected in the first place.
Several commonly used metrics are described in \cref{sec:relwork_i2g}.
However, the length of lines on a P\&ID diagram does not necessarily correlate with the actual length of pipes or other components, and thus the metrics for road network extraction is not suitable to evaluate the quality of reconstructed graphs in this domain.
In case of the engineering diagram digitization, where the connections are distinct and finite, $\text{mR@}K$ is also not suited for the digitization problem.
Thus, we propose a metric for edge detection mean Average Precision (edge mAP) using the Hungarian matching algorithm as depicted in \cref{fig:edgemetric_flow}.
The Hungarian matching algorithm is used to solve assignment problems by optimally pairing elements from two sets while minimizing the overall cost to match nodes from one graph to another using the nodes bounding boxes to use the gIoU as the cost function.
Average precision (AP) is calculated using the precision-recall curve.
A detailed description of the algorithm for edge mAP computation is \cref{alg:edgemap} in the appendix.

\begin{figure}[t] 
  \centering
  \includegraphics[trim={0 0 0cm 0cm},clip,width=0.47\textwidth]{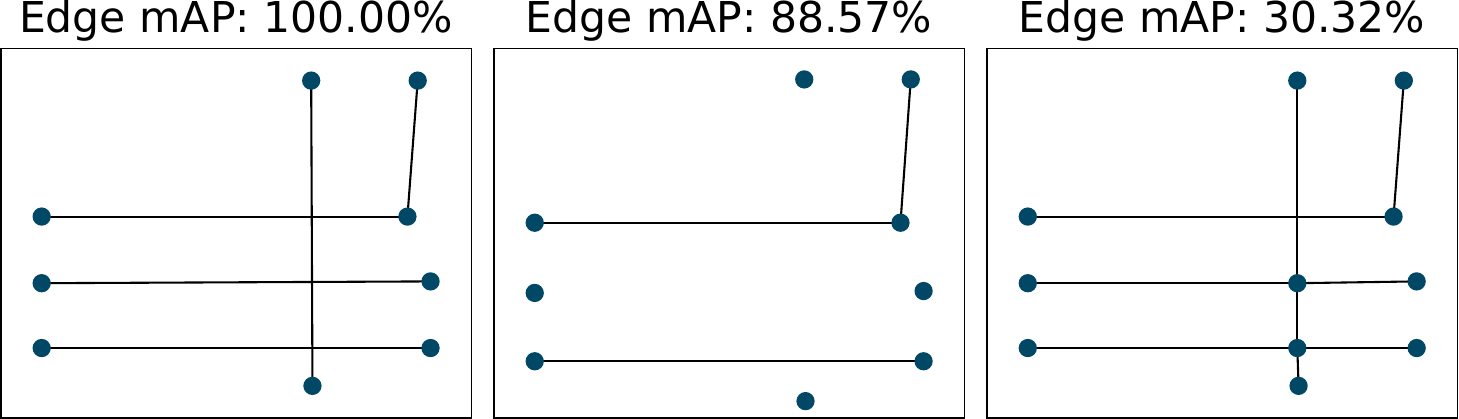}
  \caption{\label{fig:metric_example} Exemplary values of the edge mAP metric with the ground truth graph on the left and two other graphs in the middle and on the right. The graph in the middle misses two edges, and the one on the right adds several edges by falsely predicted crossing nodes.}
\end{figure}

The evaluation metric is illustrated through an example calculation in \cref{fig:metric_example}. 
This example demonstrates that the metric places greater importance on correctly predicted edges and nodes than on not predicted or wrongly predicted ones, particularly when such predictions are made adjacent to incorrectly detected crossing nodes. 

\subsection{Model Training} 

The Relationformer and Faster R-CNN are trained on a mix of the Synthetic 700 data and real world P\&IDs, as presented in \cref{tab:training_data_stats}.
First, we pre-train both methods on a large set of \num{2000} synthetic P\&IDs, which are patched and augmented multiple times to create a vast pool of \num{170944} samples for training and \num{8998} samples for validation.
Next, we fine-tune the models on a mix of the 60 real world P\&IDs and a subset of 500 synthetic P\&IDs.
To increase the presence of real data in the training set, we augment each real world P\&ID three times as much as each synthetic P\&ID.
This results in \num{44019} samples for training and \num{2317} samples for validation in the second training phase.
Approximately 37\% of the training set comprised patches from real-world data, while 63\% consisted of patches from synthetic plans.
During this second phase of training, we stop when either the loss stabilizes or the validation loss begins to rise, indicating potential overfitting.
Additional training details are listed in \cref{sec:supple_training_details}.

\begin{table}[tb]
\centering
\caption{Amount of P\&IDs and samples used for both training sets.}
\begin{tabular}{lrr}
\toprule
                        & \multicolumn{1}{c}{Pre-Training Set} & \multicolumn{1}{c}{Training Set} \\ \midrule
Training Samples   & \num{170944}                         & \num{44019}                      \\
Validation Samples & \num{8998}                           & \num{2317}                       \\
Real World P\&IDs       & 0                                    & \num{60}                         \\
Synthetic P\&IDs        & \num{2000}                           & \num{500}    \\
\bottomrule
\end{tabular}
\label{tab:training_data_stats}
\end{table}

\section{Results}

\begin{table*}[bt]
\centering
\caption{Performance comparison of the Modular Digitization and Relationformer on both test sets with higher values highlighted as bold. As in the Modular Digitzation the graph construction is done after patch merging, no values can be given for node AP and edge mAP for patches.}
\setlength{\tabcolsep}{5pt}
\begin{tabular}{lclll@{\hskip 12pt}clll@{\hskip 17pt}clll@{\hskip 12pt}cll}
\toprule
                & \multicolumn{7}{c}{Modular Digitization}                      &  & \multicolumn{7}{c}{Relationformer}                            \\
                & \multicolumn{3}{c}{Patched} &  & \multicolumn{3}{c}{Stitched} &  & \multicolumn{3}{c}{Patched} &  & \multicolumn{3}{c}{Stitched} \\
 &
  Symbols &
  \multicolumn{1}{c}{Nodes} &
  \multicolumn{1}{c}{Edges} &
   &
  Symbols &
  \multicolumn{1}{c}{Nodes} &
  \multicolumn{1}{c}{Edges} &
   &
  Symbols &
  \multicolumn{1}{c}{Nodes} &
  \multicolumn{1}{c}{Edges} &
   &
  Symbols &
  \multicolumn{1}{c}{Nodes} &
  \multicolumn{1}{c}{Edges} \\
 &
  mAP &
  \multicolumn{1}{c}{AP} &
  \multicolumn{1}{c}{mAP} &
   &
  mAP &
  \multicolumn{1}{c}{AP} &
  \multicolumn{1}{c}{mAP} &
   &
  mAP &
  \multicolumn{1}{c}{AP} &
  \multicolumn{1}{c}{mAP} &
   &
  mAP &
  \multicolumn{1}{c}{AP} &
  \multicolumn{1}{c}{mAP} \\ \midrule
PID2Graph OPEN100& \multicolumn{1}{l}{\textbf{86.58}} & -     & -    & & \multicolumn{1}{l}{\textbf{76.99}}   & 52.14 & 45.89 & & \multicolumn{1}{l}{73.49} & 82.18 & 76.79 & & \multicolumn{1}{l}{73.14}  & \textbf{83.63} & \textbf{75.46} \\
PID2Graph Synthetic& \multicolumn{1}{l}{\textbf{78.74}}           & -     & -     &  & \multicolumn{1}{l}{74.15}           & 85.16          & 50.26         &  & \multicolumn{1}{l}{78.62}  & 87.44 & 93.86 &  & \multicolumn{1}{l}{\textbf{78.36}}  & \textbf{96.89} & \textbf{88.95} \\
Dataset-P\&ID & \multicolumn{1}{l}{\textbf{85.87}} & -     & -    & & \multicolumn{1}{l}{\textbf{83.95}}   & 89.32 & 85.46 & & \multicolumn{1}{l}{80.32} & 96.59 & 92.46 & & \multicolumn{1}{l}{76.69}  & \textbf{97.72} & \textbf{95.07} \\ \bottomrule
\end{tabular}
\label{tab:results_1}
\end{table*}

The results are presented in \cref{tab:results_1}.
Our experiments on the patched OPEN100 data with the Relationformer model demonstrate good performance in detecting symbols (73.49\%), nodes (82.18\%) and connections (76.79\%).
When evaluated on full plans, the performance stays the same (around 1\% increase or decrease).
For the synthetic data, the symbol detection mAP has similar values after stitching the patches, while the stitching has a significant influence on node and edge detection, rising the node AP by 9.45\% to 96.89\% and lowering the edge mAP by 4.91\% to 88.95\%.
In contrast, the modular digitization shows a more moderate performance, achieving decent symbol detection on the OPEN100 data on patches (86.58\%) but struggling with node detection (52.15\%) and edge detection (45.89\%), similar to its performance on synthetic data.

Both methods demonstrate strong results on \mbox{Dataset-P\&ID}, with modular digitization again showing better values only for symbol detection.

To investigate the impact of larger symbols, we conducted experiments on the OPEN100 data using an enlarged patch size to encompass bigger symbols within a single patch. 
Additionally, we performed experiments on a subset of the OPEN100 data that consisted only of plans where all symbols could be accommodated entirely within a patch.
The results can be seen in \cref{tab:results_ablation}.
When increasing patch size from originally (1500, 1500) to (2000, 2000), which should facilitate better inclusion of larger symbols, the symbol and node detection value of the Relationformer drop by at least 10\%, while the values of the Modular Digitization stay around the same.
When using the small symbol subset, both methods show improved performance, with a gain of 9.49\% for symbol detection by the Relationformer and 10.49\% by the Modular Digitization.
Moreover, the results show that the Modular Digitization's performance drops significantly from patched to stitched on the full OPEN100 data. 
However, its performance increases when using only the subset with small symbols. 
This suggests that the stitching process struggles with larger symbols.

\begin{table}[tb]
\centering
\caption{Ablation studies for the Relationformer and Modular Digitization for the OPEN100 data on different patch sizes and a test subset containing only objects that fit into a patch completely.}
\begin{tabular}{@{}llllll@{}}
\toprule
& \multirow{2}{*}{Data} & \multicolumn{1}{c}{\multirow{2}{*}{Patch Size}} & \multicolumn{1}{c}{Symbols} & \multicolumn{1}{c}{Nodes} & \multicolumn{1}{c}{Edges} \\
& \multirow{2}{*}{} & \multicolumn{1}{c}{} & \multicolumn{1}{c}{mAP} & \multicolumn{1}{c}{AP} & \multicolumn{1}{c}{AP} \\ 
\midrule
 & Patched & (2000, 2000) & 86.52 & - & - \\
\multirow{4}{*}{\rotatebox[origin=c]{90}{\parbox[c]{1cm}{\centering \,\,\,\,Modular \\ Digitization}}} & Stitched & (2000, 2000) & 76.74 & 57.49 & 55.08 \\
 & \renewcommand{\arraystretch}{0.3}\begin{tabular}[c]{@{}l@{}}Patched, Small \\ Symbol Subset\end{tabular} & (1500, 1500) & 86.51 & - & - \\
& \renewcommand{\arraystretch}{0.3}\begin{tabular}[c]{@{}l@{}}Stitched, Small \\ Symbol Subset\end{tabular} & (1500, 1500) & 87.48 & 62.26 & 53.95 \\
 \midrule
& Patched & (2000, 2000) & 62.23 & 50.11 & 69.88 \\
\multirow{4}{*}{\rotatebox[origin=c]{90}{\parbox[c]{0.65cm}{\centering Relation$-$ \\ \,\,\,former}}} & Stitched & (2000, 2000) & 62.42 & 56.15 & 67.67 \\
& \renewcommand{\arraystretch}{0.3}\begin{tabular}[c]{@{}l@{}}Patched, Small \\ Symbol Subset\end{tabular} & (1500, 1500) & 80.07 & 82.90 & 73.08 \\
& \renewcommand{\arraystretch}{0.3}\begin{tabular}[c]{@{}l@{}}Stitched, Small \\ Symbol Subset\end{tabular} & (1500, 1500) & 82.74 & 83.80 & 75.16 \\
 \bottomrule
\end{tabular}
\label{tab:results_ablation}
\end{table}

The importance of real-world data in the training and test datasets is further underscored by an ablation study presented in \cref{sec:supple_ablation}, which demonstrates that the Relationformer's performance suffers significantly when trained on synthetic data only, particularly when evaluated on the OPEN100 dataset.

\begin{figure}[tb] 
  \centering
  \includegraphics[trim={0, 0, 0, 0},clip,width=.44\textwidth] {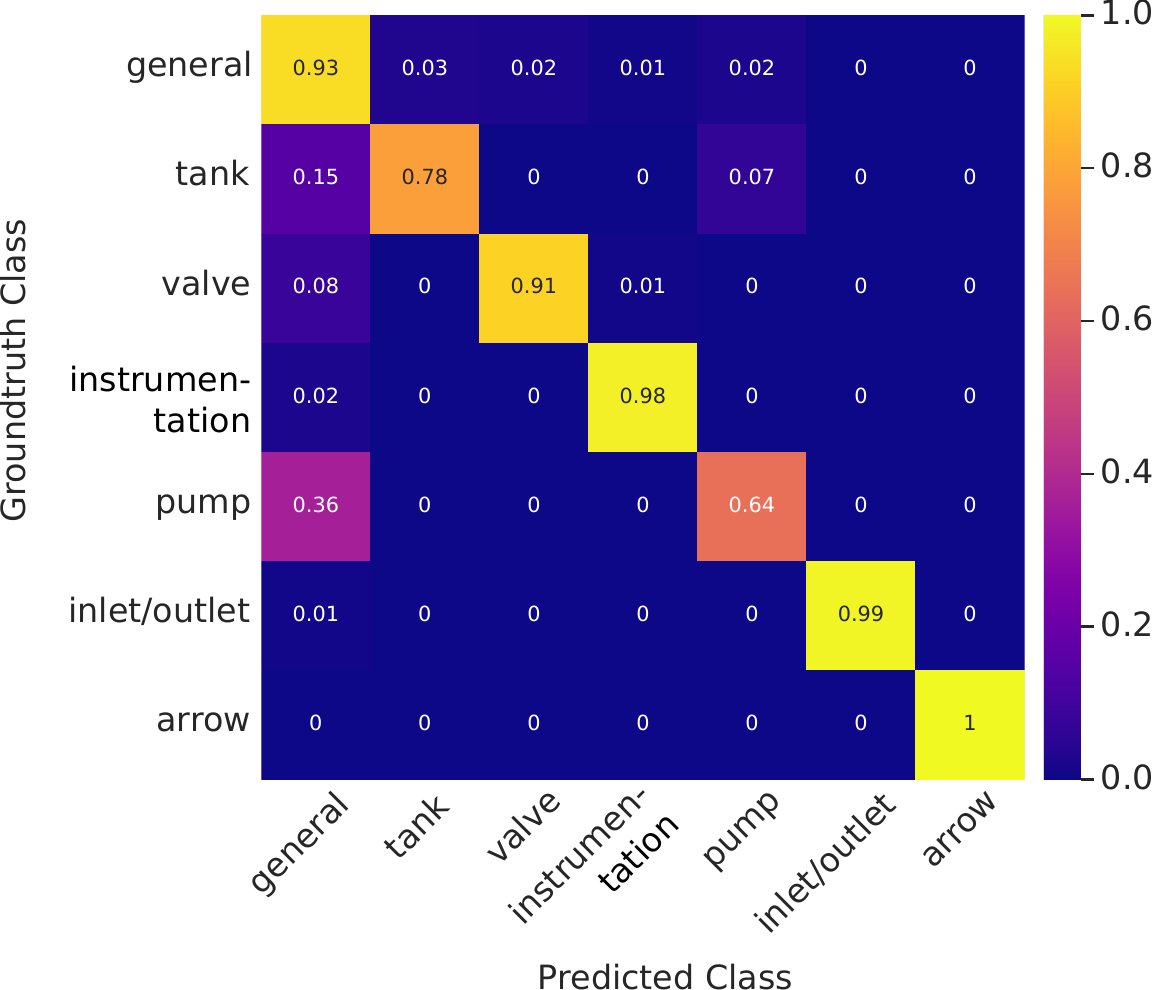}
  \caption{\label{fig:conf_matrix_1} Confusion matrix for the stiched symbol detection results of the Relationformer for the OPEN100 data.}
\end{figure}

The performance of the Relationformer on the OPEN100 data is further investigated by examining the confusion matrix in \cref{fig:conf_matrix_1}. 
Notably, the 'general' class exhibits a high degree of confusion, with pumps being frequently misclassified as symbols belonging to this category.
The confusion with the 'general' class is consistent with the other datasets (see \cref{sec:supplementary_general_class} for the remaining confusion matrices).

\begin{figure}[bt] 
  \centering
  \includegraphics[trim={2cm 4cm 6.5cm 0cm},clip,width=.47\textwidth] {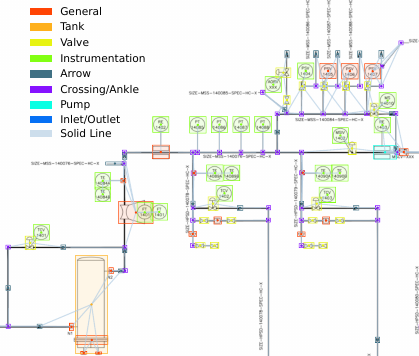}
  \caption{\label{fig:prediction_2} Extract of a digitization result of the Relationformer for one merged P\&ID from the OPEN100 test data. The legend shows the class names of the predictions.}
\end{figure}

To gain a better understanding of the detection capabilities and graph reconstruction quality, we have also visually inspected the predictions.
\cref{fig:prediction_2} illustrates the merged graph for an entire plan.
Overall, the graph is correctly constructed, though there are some noticeable misclassifications of instrumentation symbols and additional false positive diagonal connections between crossings and symbols.

\section{Discussion}
\label{sec:discussion}

The Relationformer shows good results on every task, while the Modular Digitization shows especially bad results on the edge detection.
The results highlight several challenges that need to be addressed in order to achieve accurate graph reconstruction and relation detection.
The symbol detection module of the Modular Digitization however achieves better performance in correctly classifying symbols on the OPEN100 data.

The Modular Digitization has the disadvantage of being highly dependent on previous steps, where each step, especially line detection, is reliant on using good parameters.
The Relationformer enables end-to-end training without significant adjustment after training.
This property is expected to facilitate good generalization across different domains.

One of the primary difficulties lies in the patching and merging process.
Errors occurred during the patching process itself, resulting in incorrect input data for our models used for training.
When splitting the P\&ID diagrams into patches, this can lead to the truncation of symbols. 
When only a fragment of a symbol remains, it can be difficult to distinguish it from other symbols or even lines, making it difficult for our models to train and evaluate on this data.

A problem with detecting symbols and differentiating between classes is the "general" class.
This class encompasses a diverse range of symbols, resulting in high intra-class variation.
As a consequence, models struggle to generalize effectively across these disparate symbols, making evaluation and pinpointing specific issues within this category more challenging.
The confusion matrices of the Relationformer support this assertion, as the general class is the class with the highest confusion across all datasets.
The general detection and localization of symbols appear to be quite effective, as indicated by high node AP scores. 
However, improving the classification of symbols not seen during training and assigning more specific labels remains a focus for future work.

The metrics applied are based on classical symbol detection techniques and include a custom-defined metric. We used an IoU threshold of 0.5 to evaluate our methods, which is relatively low compared to other object detection tasks. 
A potential challenge in using a higher IoU threshold lies in the uncertainty of manually annotated data, as these annotations are less precise than the automatically generated bounding boxes for synthetic data. 
The utility of the edge metric warrants consideration, since it disproportionately penalizes incorrect edge predictions compared to missed edges.
Furthermore, the edge detection metric relies on accurately matching nodes between the ground truth and predicted graphs based on bounding boxes.
However, if a method accurately predicts edges, the metric will reflect this performance appropriately.

Additionally, our study underscores the importance of collecting and utilizing larger datasets with real-world diagrams including big symbols.
Even though dashed or non-continuous lines are not present in the OPEN100 data, the results on the synthetic data suggests that the Relationformer is also able to classify line types accurately.

\section{Conclusion}
Leveraging the state-of-the-art Relationformer architecture, we propose a pipeline that simultaneously detects objects and their relationships from engineering diagrams.
In this study, we address the challenge of digitizing engineering diagrams of complex technical systems, specifically Piping and Instrumentation Diagrams (P\&IDs).
By adapting the Relationformer architecture, we develop a robust pipeline that can simultaneously identify objects and their relationships within these diagrams. 
To facilitate evaluation and comparison, we introduce a novel, publicly available dataset and establish a set of meaningful metrics. 
Our approach yields impressive results on both real-world and synthetic P\&IDs, outperforming a modular digitization approach.
Notably, our method achieves an AP of 83.63\% for node detection and an edge mAP of 75.46\%, highlighting its effectiveness in accurately extracting valuable information from complex engineering diagrams.

\section*{Acknowledgment}
We acknowledge the assistance of Llama 3.2 \cite{llama32} throughout all sections of the paper to enhance the readability and clarity by improving formulations and overall English.
The authors would like to thank Enrique Ríos Smits for his assistance in processing and labeling the data used in this work.

\bibliographystyle{IEEEtranN}
\bibliography{IEEEabrv,literature}

\clearpage

\section*{APPENDIX}
\setcounter{equation}{0}
\setcounter{figure}{0}
\setcounter{table}{0}
\setcounter{section}{0}

\renewcommand{\thetable}{A\arabic{table}}
\renewcommand{\thefigure}{A\arabic{figure}}
\renewcommand{\thealgorithm}{A\arabic{algorithm}}
\renewcommand{\thesection}{A\arabic{section}}


\section{Edge Metric Calculation}

We calculate the metric for edge detection mean Average Precision (mAP) as described in \cref{alg:edgemap}.
The Hungarian matching algorithm \cite{Kuhn.1955} is used to match nodes (mapping $M$) from the ground truth graph $G_\text{true}$ to the predicted graph $G_\text{pred}$ using the nodes (denoted with $V$) bounding boxes.
All edges $e_{u,v}$ from node $u$ to node $v$ of the predicted graph have a confidence score and a class.
An edge $e_{\hat{u},\hat{v}}$ is an edge in the ground truth graph and also has a class.
The Python library \texttt{scikit-learn} \cite{scikit-learn} is used to calculate the average precision.

\begin{algorithm}[h]
    \caption{Edge mAP Computation}
    \begin{algorithmic}[1]
        \STATE \textbf{Input:} $G_{\text{true}}(V_{\text{true}}, E_{\text{true}})$, $G_{\text{pred}}(V_{\text{pred}}, E_{\text{pred}})$
        \STATE \textbf{Initialize:} 
        \STATE $(\text{TP, FP, FN}) \gets (\text{empty list, empty list, empty list})$
        
        \STATE $M \gets$ HungarianMatcher($V_{\text{true}}, V_{\text{pred}}$) $:V_{\text{true}} \to V_{\text{pred}}$
        
        \FOR{each predicted edge $e_{u, v} \in E_{\text{pred}}$}
            \STATE $e_{\hat{u},\hat{v}}$ $\gets$ $e_{M^{-1}(u), M^{-1}(v)}$
            \IF{$e_{\hat{u},\hat{v}} \in E_{\text{true}}$}
                \IF{class($e_{u,v}$) = class($e_{\hat{u},\hat{v}}$)}
                    \STATE TP.insert($(\text{class}(e_{u,v}), \text{conf\_score}(e_{u,v}))$)
                \ELSE
                    \STATE FP.insert($(\text{class}(e_{u,v}), \text{conf\_score}(e_{u,v}))$)
                \ENDIF
            \ELSE
                \STATE FP.insert($(\text{class}(e_{u,v}), \text{conf\_score}(e_{u,v}))$)
            \ENDIF
        \ENDFOR
        
        \FOR{each true edge $e_{\hat{u}, \hat{v}} \in E_{\text{true}}$}
            \STATE $e_{u,v} \gets e_{M(\hat{u}), M(\hat{v})}$
            \IF{$e_{u,v} \notin E_{\text{pred}}$}
                \STATE FN.insert($\text{class}(e_{\hat{u},\hat{v}})$)
            \ENDIF
        \ENDFOR
        
        \STATE $\text{mAP}_{\text{edge}} \gets \frac{1}{|\text{classes}|} \sum_{c \in \text{classes}} \text{AP}_c (\text{TP, FP, FN})$
        
        \STATE \textbf{Return:} $\text{mAP}_{\text{edge}}$
    \end{algorithmic}
    \label{alg:edgemap}
\end{algorithm}

\section{Training}
\label{sec:supple_training_details}

All training, testing, and experimental procedures were conducted on a single NVIDIA Quadro RTX 8000 graphics card with 48GB of memory.
The relevant hyperparameters can be seen in \cref{tab:relationformer_training_settings} and \cref{tab:fasterrcnn_training_settings}.
For both Relationformer and Faster R-CNN with the Modular Digitization, we employ a 0.95:0.05 training-validation split, allocating the majority of the data to training. 
This partitioning is deliberate, given the transformer network's propensity for requiring substantial amounts of data to achieve good performance.

\begin{table}[ht]
\centering
\caption{Hyperparameters used to train the Relationformer.}
\label{tab:relationformer_training_settings}
\begin{tabular}{ll}
\toprule
Hyperparameter            & Value      \\ \midrule
Batch Size                & 20         \\
Image Size                & (512, 512) \\
Epochs                    & 80         \\
Learning Rate             & 1e-4       \\
Learning Rate Backbone    & 3e-5       \\
Number of Queries         & 401        \\
Backbone                  & ResNet-101 \\
$\lambda_{\text{box}}$, $\lambda_{\text{cls}}$, $\lambda_{\text{card}}$, $\lambda_{\text{node}}$, $\lambda_{\text{edge}}$ & 2, 2, 1, 4, 3 \\
Randomize Edge Directions & Yes       \\
\bottomrule
\end{tabular}
\end{table}

\begin{table}[ht]
\centering
\caption{Hyperparameters used to train Faster R-CNN for symbol detection.}
\label{tab:fasterrcnn_training_settings}
\begin{tabular}{ll}
\toprule
Hyperparameter            & Value      \\ \midrule
Batch Size                & 16         \\
Image Size                & (1024, 1024) \\
Epochs                    & 40         \\
Learning Rate             & 5e-4       \\ 
\bottomrule
\end{tabular}
\end{table}

\subsection{Dataset Details}
\label{sec:relwork_dataset_details}

The datasets analyzed in this study prior to patching are summarized in \cref{tab:data_details}, which reveals key characteristics.
Specifically, the table reports the total number of symbols and edges, mean values for node and edge counts per plan, as well as the size distribution of symbols.
Notably, real-world P\&IDs exhibit a tendency towards increased complexity, characterized by higher mean numbers of nodes and edges per plan, as well as greater variance compared to synthetic data.

\begin{table*}[t]
\centering
\caption{Dataset statistics prior to splitting and augmentation. Real-world P\&IDs exhibit significantly higher data variability compared to synthetic counterparts.}
\begin{tabular}{@{}lllrrll@{}}
\toprule
& \begin{tabular}[c]{@{}l@{}}\# of Nodes \\ per Plan \end{tabular} & \begin{tabular}[c]{@{}l@{}}\# of Edges \\ per Plan \end{tabular} & \begin{tabular}[c]{@{}l@{}}\# of Symbols, \\ Crossings \\ \& Ankles \end{tabular} & \# of Edges & \begin{tabular}[c]{@{}l@{}}Symbol Area (px) \\ Mean ± Std \end{tabular} & \begin{tabular}[c]{@{}l@{}}Usage\end{tabular} \\ \midrule
Synthetic 700 & \num{131} ± \num{14} & \num{133} ± \num{15} & \num{183901}   & \num{188550} & \num{6071} ± \num{22059} & Train\\
Real World   & \num{497} ± \num{338} & \num{457} ± \num{338} & \num{17218}  & \num{14649} & \num{5776} ± \num{54328} & Train \\
Dataset-P\&ID  & \num{452} ± \num{86} & \num{505} ± \num{91} & \num{51182} & \num{137449} & \num{3350} ± \num{4291} & Test \\
\begin{tabular}[c]{@{}l@{}}PID2Graph Synthetic \end{tabular} & \num{131} ± \num{14} & \num{133} ± \num{15} & \num{23323} & \num{23994} & \num{4208} ± \num{8993} & Test \\
PID2Graph OPEN100             & \num{382} ± \num{125} & \num{381} ± \num{126} & \num{2173}    & \num{2100} & \num{13644} ± \num{69234} & Test \\
PID2Graph OPEN100 Subset      & \num{414} ± \num{44} & \num{412} ± \num{47} & \num{793}    & \num{765} & \num{12442} ± \num{60297} & Test \\ \bottomrule
\end{tabular}
\label{tab:data_details}
\end{table*}

\section{Further Analysis of General Class}
\label{sec:supplementary_general_class}

\begin{figure}[b]%
    \centering
    \subfloat[\centering]{{\includegraphics[width=0.23\textwidth]{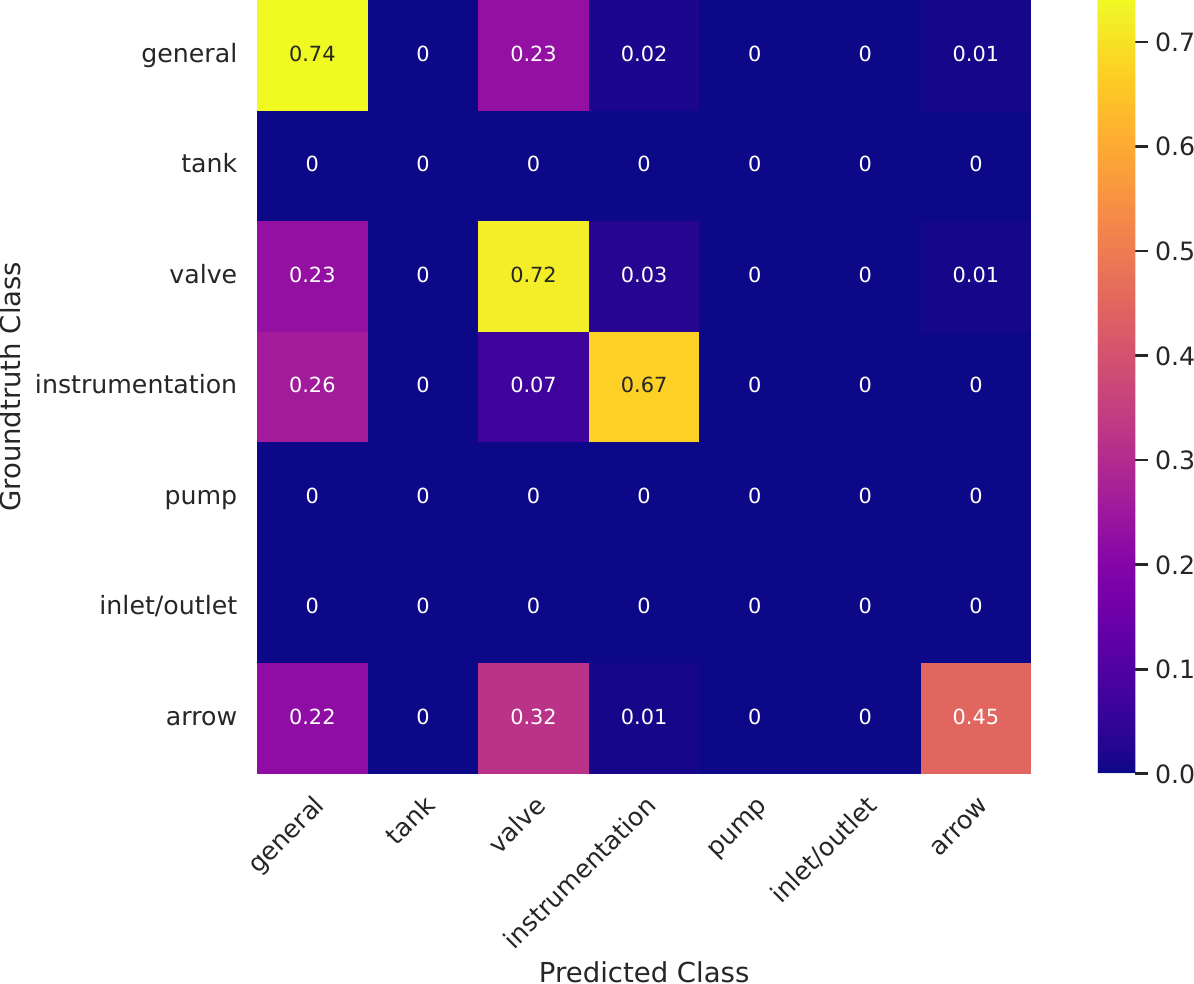}}}%
    \hfill%
    \subfloat[\centering]{{\includegraphics[width=0.23\textwidth]{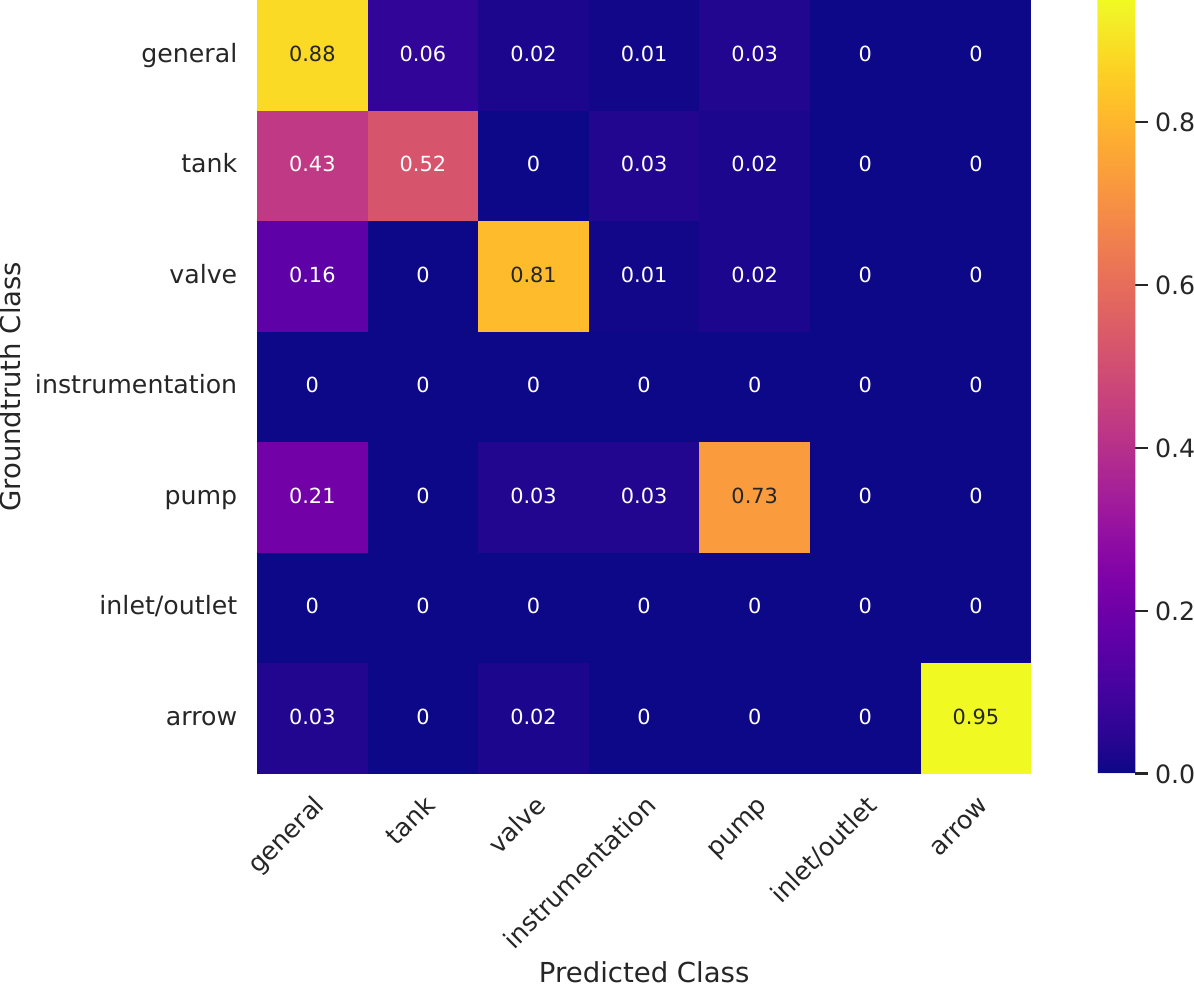}}}%
    \caption{Confusion matrices for stitched symbol detection results of the Relationformer for (a) Dataset-P\&ID and (b) PID2Graph Synthetic.}%
    \label{fig:conf_matrix_2}%
\end{figure}

The confusion matrix of the Relationformer on the PID2Graph OPEN100 dataset is shown in \cref{fig:conf_matrix_1}, while the confusion matrices of the Relationformer on the PID2Graph Synthetic and Dataset-P\&ID are shown in \cref{fig:conf_matrix_2}.

The confusion is the highest with the 'general' class for PID2Graph and Dataset-P\&ID, with confusion ranging up to 43\%.

\begin{figure*}[bt]%
    \centering
    \subfloat[\centering]{{\includegraphics[width=0.44\textwidth]{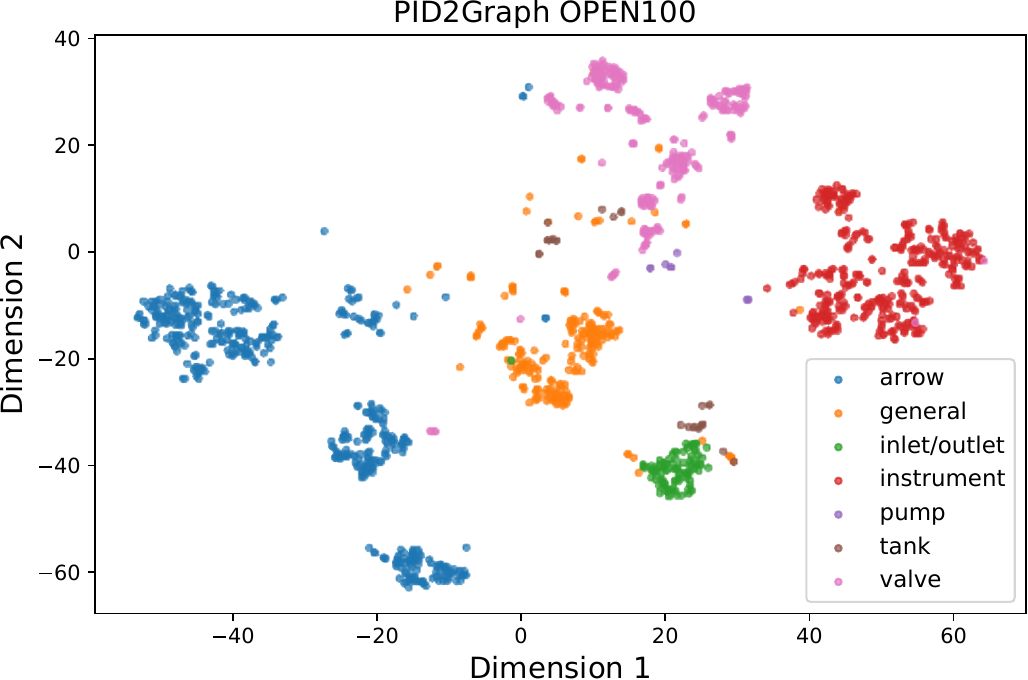}}}%
    \hfil%
    \subfloat[\centering]{{\includegraphics[width=0.44\textwidth]{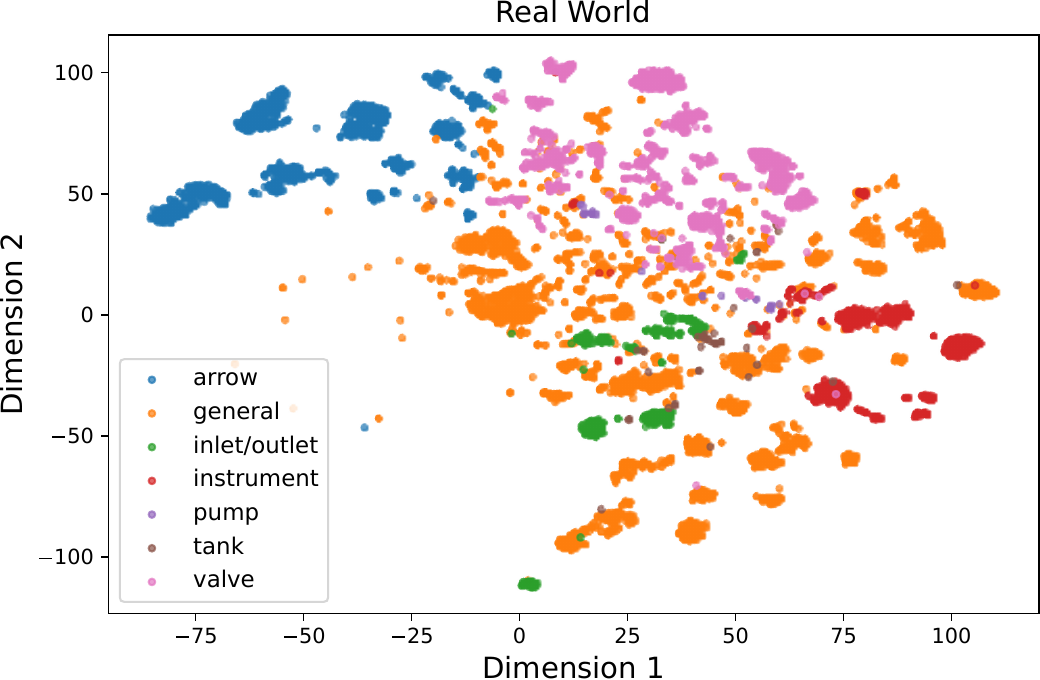}}}%
    \caption{2D t-SNE visualization of features by class for (a) PID2Graph OPEN100 and (b) Real World.}%
    \label{fig:tsne}%
\end{figure*}

To gain deeper insight into the diversity within each class, we extracted feature vectors from every symbol in our dataset using ResNet-101 with an input size of (224, 224).
We then applied t-SNE to visualize these features in a 2D space (see \cref{fig:tsne}). 
This analysis reveals that the 'general' class exhibits the highest degree of centralization and overlap with other classes across all datasets.
This also suggests substantial intra-class variability for the 'general' category.

\section{Dataset-P\&ID Comparison}

The line detection precision of DigitizePID is not directly comparable to edge mAP, however we display both of them.

Our evaluation methods required converting the Dataset-P\&ID \cite{Paliwal.2021} data into a graph structure (as described in Section 3.4). 
Notably, we used an Intersection-over-Union (IoU) threshold of 75\% for calculating mAP values, which is higher than the 50\% threshold used in our other experiments.
The line detection precision of DigitizePID is not directly comparable to our edge mAP results, but both metrics are presented for reference. 

\begin{table}[tb]
\centering
\caption{Performance comparison of the Modular Digitization, the Relationformer and DigitizePID on the synthetic Dataset-P\&ID dataset.}
\begin{tabular}{cllll}
\toprule
& \multicolumn{4}{c}{Dataset-P\&ID \cite{Paliwal.2021}, Stitched} \\
 & Symbols & \multicolumn{1}{c}{Nodes} & \multicolumn{1}{c}{Edges} & \multicolumn{1}{c}{Lines} \\
 & mAP & \multicolumn{1}{c}{AP} & \multicolumn{1}{c}{mAP} & Accuracy \\ \midrule
Modular Digitization & \multicolumn{1}{l}{69.82}  & 85.42 & 85.46 & - \\
Relationformer       & \multicolumn{1}{l}{43.04}  & 75.85 & 95.07 & - \\
\multicolumn{1}{c}{\begin{tabular}[c]{@{}c@{}}DigitizePID \cite{Paliwal.2021}\end{tabular}} & \multicolumn{1}{l}{92.50}  & - & - & 91.13 \\  \bottomrule
\end{tabular}
\label{tab:results_dpid}
\end{table}

 Despite this, the results show that both Modular Digitization and Relationformer perform poorly compared to DigitizePID, likely due to differences in IoU thresholds and bounding box styles. 
 Specifically, the loose-fitting bounding boxes in Dataset-P\&ID, combined with smaller symbol sizes (cf. \cref{sec:relwork_dataset_details}), resulted in non-detections at this higher IoU. 
 However, our results indicate that the Relationformer's edge detection is robust and effective.

\section{Ablation}
\label{sec:supple_ablation}

To underscore the value of training on real-world data, we conducted a comparative evaluation of the Relationformer model, examining its performance when trained exclusively on synthetic P\&ID data versus when trained on a combination of synthetic data and the mix of synthetic and real-world data.
The results in \cref{tab:supple_synth_training} show a consistent improvement in performance when the model is trained on the mixed dataset.
Notably, the largest performance gap is observed when comparing the results on the OPEN100 dataset, highlighting the benefits of incorporating diverse and realistic training data.

\begin{table}[bt]
\centering
\caption{Relationformer performance comparison when trained on synthetic data only versus the mix of synthetic and real-world data.}
\setlength{\tabcolsep}{4pt}
\label{tab:supple_synth_training}
\begin{tabular}{@{}ccllll@{}}
\toprule
\multicolumn{2}{c}{Trained on}   & \multicolumn{1}{c}{\multirow{3}{*}{Test Dataset}} & \multicolumn{3}{c}{Patched} \\
\multicolumn{1}{l}{\multirow{2}{*}{Synthetic}} & \multicolumn{1}{c}{\multirow{2}{*}{Mix}} &                                  & Symbols   & Nodes  & Edges    \\
\multicolumn{1}{l}{}                           & \multicolumn{1}{c}{}                     &                                  & mAP       & AP     & AP      \\ \midrule
\checkmark                       &                                                 & PID2GRAPH OPEN100                & 26.67     & 34.35  & 44.49   \\
\checkmark                       &                                                 & PID2GRAPH Synthetic              & 63.20     & 73.71  & 88.02   \\
\checkmark                       &                                                 & Dataset-P\&ID                    & 36.36     & 72.45  & 84.88   \\ \midrule
\checkmark                       & \checkmark                       & PID2GRAPH OPEN100                & 73.49     & 82.18  & 76.79   \\
\checkmark                       & \checkmark                       & PID2GRAPH Synthetic              & 78.62     & 87.44  & 93.86   \\
\checkmark                       & \checkmark                       & Dataset-P\&ID                    & 80.32     & 96.59  & 92.46   \\ \bottomrule
\end{tabular}
\end{table}

\end{document}